%% file: top.tex
\pdfoutput=1
\documentclass[10pt,twocolumn,letterpaper]{article}

\usepackage{iccv}
\usepackage{times}
\usepackage{epsfig}
\usepackage{graphicx}
\usepackage{amsmath}
\usepackage{amssymb}
\usepackage{multirow}
\input{defs}

\usepackage[pagebackref=true,breaklinks=true,letterpaper=true,colorlinks,bookmarks=false]{hyperref}

 \iccvfinalcopy 

\def\iccvPaperID{1844} 

\begin{document}

\title{Globally Optimal Cell Tracking using Integer Programming}

\author{Engin~T\"{u}retken\thanks{\:  The authors contributed equally.}\\
CSEM, Switzerland\\
{\tt\small engin.tueretken@alumni.epfl.ch}
\and
Xinchao Wang$^{*}$ \\
CVLab, EPFL, Switzerland\\
{\tt\small xinchao.wang@epfl.ch}
\and
Carlos J. Becker\\
CVLab, EPFL, Switzerland\\
{\tt\small carlos.becker@epfl.ch}
\and
Carsten Haubold\\
HCI, University of Heidelberg, Germany\\
{\tt\small carsten.haubold@iwr.uni-heidelberg.de }
\and
Pascal Fua\\
CVLab, EPFL, Switzerland\\
{\tt\small pascal.fua@epfl.ch}
}

\maketitle

\input{abstract}

\input{intro}

\input{related}

\input{method}

\input{exp}

\input{conclusion}

{\small
\bibliographystyle{ieee}
\bibliography{short,vision,learning,biomed,misc,optim,graphics,photog}
}

\end{document}

%% file: defs.tex
\usepackage{color}
\usepackage{subfig}

\newcommand{\comment}[1]{}
\newcommand{\note}[1]{{\color{red}{\noindent\framebox{\begin{minipage}{8cm}{#1}\end{minipage}}\\ }}}

\usepackage{array}
\newcolumntype{C}{>{\centering\arraybackslash} m{0.32\linewidth} }

\newcommand{\gmm}[0]{{\bf  GMM}}
\newcommand{\kth}[0]{{\bf  KTH}}
\newcommand{\ct}[0]{{\bf   CT}}
\newcommand{\jst}[0]{{\bf   JST}}

\newcommand{\ours}[0]{{\bf OURS}}
\newcommand{\cl}[0]{{\bf OURS-CL}}
\newcommand{\bh}[0]{{\bf OURS-BH}}
\newcommand{\lp}[0]{{\bf OURS-LP}}
\newcommand{\nc}[0]{{\bf OURS-NC}}
\newcommand{\fd}[0]{{\bf OURS-FD}}

\newenvironment{packed_item}{
\begin{itemize}
  \setlength{\itemsep}{1pt}
  \setlength{\parskip}{0pt}
  \setlength{\parsep}{0pt}
}{\end{itemize}}

%% file: abstract.tex

\begin{abstract}

We  propose a  novel  approach  to automatically  tracking  cell populations  in
time-lapse images. 
To account for cell occlusions and overlaps, 
we introduce a robust method that 
generates an over-complete set of competing detection hypotheses.
We then  perform  detection and  tracking simultaneously on these hypotheses 
by solving to optimality an integer program with only one type of flow variables.
This eliminates the need for heuristics to  handle missed detections  
due  to  occlusions  and  complex  morphology.   
We  demonstrate  the effectiveness  of  our  approach  on  a range  of  
challenging sequences consisting  of  clumped cells  and show  that it  
outperforms  state-of-the-art techniques.

\end{abstract}

%% file: intro.tex

\section{Introduction}
\label{sec:intro}

\input{figs/motivation.tex}

Detecting  and  tracking  cells  over  time is  key  to  understanding  cellular
processes  including  division  (mitosis),  migration,  and  death  (apoptosis).
Modern microscopes produce vast image streams making manual tracking tedious and
impractical.  High-throughput  automated systems  are therefore  increasingly in
demand and  several cell tracking  competitions have recently been  organized to
attract    Computer    Vision    researchers'     interest    and    speed    up
progress~\cite{Maska14,Solorzano14}.    These  competitions   have  shown   that
state-of-the-art methods are still error-prone due to occlusions, imaging noise,
and complex cell morphology.

Cell tracking is  an instance of the more generic  multi-target tracking problem
with the additional difficulties that cells, unlike for example pedestrians, can
either divide or wither away and disappear in mid-sequence. In its generic form,
the  problem is  often  formulated as  a two-step  process  that involves  first
detecting  potential  objects  in  individual  frames  and  then  linking  these
detections  into complete  trajectories.   This approach  is attractive  because
spurious detections, such as false positives due to imaging noise and artifacts,
can be eliminated by imposing temporal  consistency across many frames, which is
more difficult to do in recursive approaches.

Many      of      the      most     successful      algorithms      to      both
people~\cite{Ge08,Andriyenko12,Yang12a,Wojek13}             and             cell
tracking~\cite{Li08b,Kausler12,Schiegg13} follow this two-step approach by first
running an  object detector  on each  frame independently  and building  a graph
whose nodes are the detections and edges connect pairs of them. They then find a
subgraph that represents  object trajectories by considering the  whole graph at
once.   However,  to handle  missed  detections,  these  methods often  rely  on
heuristic  procedures  that  offer  no  guarantee of  optimality.   This  is  of
particular concern for cell tracking because detections are often unreliable due
to   the   complex   morphology   of  cell   populations.    For   example,   in
Fig.~\ref{fig:motivation}, groups of cells that  appear clumped together in some
frames can only  be told apart when  considering the sequence as  a whole.

The   approach  proposed  in~\cite{Schiegg14}  addresses   this  issue  by
  introducing a factor graph for joint cell segmentation and tracking. Inference
  is then achieved by solving a relatively complex integer program that involves
  several  types of  variables  and  constraints. The  algorithm  starts from  a
  watershed-based  oversegmentation that  is   sensitive  to inaccuracies in  
  pixel probability estimates. It is therefore subject to errors when the cells
  are clumped together and hard to distinguish from each other.

In this paper, we formulate joint detection and tracking in terms of constrained
Bayesian  inference.  This  lets  us  cast the  problem  as  an integer  program
expressed in  terms of a single  type of flow variables.   To explicitly account
for  the  often  ambiguous  output  of  cell  detectors,  we  devised  a  robust
ellipse-fitting  algorithm  to  generate multiple  and  potentially  conflicting
initial  hypotheses  from the  output  of  a foreground-background  segmentation
algorithm, as depicted by  Fig.~\ref{fig:close_up_hierarchy}.  We then model critical
cell events,  such as migration and  division, using flow variables  and resolve
the conflicts by  solving the resulting integer program, which  is   conceptually  
much   simpler  than   those  of   earlier
approaches~\cite{Kausler12,Schiegg13,Jug14,Schiegg14}.      It    also
features  fewer   variables  and  constraints,  which   implies  better  scaling
properties. 

Although network flow formulation has been used in this context before, existing
approaches~\cite{Padfield09b,Padfield11} focus on cell  tracking in consecutive image pairs,
while  our approach  aggregates  image  evidences from  the  whole sequence  and
conducts  global optimization  over all  potential  cell locations  at all  time
frames.   Furthermore, we  handle competing  hypotheses within  the same optimization
framework instead of having to decide at detection time.

We  show  that   this  improves  trajectories  and   yields  superior  detection
performance   on   various   datasets   compared  to   the   recent   approaches
of~\cite{Schiegg13,Amat14,Magnusson14,Schiegg14},  which includes  the technique
that     performed     best     on     the     above-mentioned     cell-tracking
challenges~\cite{Maska14,Solorzano14}.

\input{figs/ellipses.tex}



%% file: figs/motivation.tex
\begin{figure*}[!t]
\begin{center}
\hspace{-7pt}
\begin{tabular}{cccccccc}
   & \hspace{-0.3cm}\normalsize{Source Images} 
   & \normalsize{\hspace{-10pt} Segmentation}  
   & \normalsize{\hspace{-10pt} \ct{}~\cite{Schiegg13}} 
   & \normalsize{\hspace{-10pt} \jst{}~\cite{Schiegg14}}
   & \normalsize{\hspace{-10pt} \kth{}~\cite{Magnusson12}} 
   & \normalsize{\hspace{-10pt} \ours{}} 
   & \normalsize{\hspace{-15pt} Ground Truth}\\
    \hspace{-0.0cm}\rotatebox{90}{\hspace{0.75cm} t=76}&  
   \hspace{-0.3cm}\includegraphics[width=0.125\linewidth]{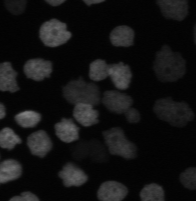} &
   \hspace{-0.3cm}\includegraphics[width=0.125\linewidth]{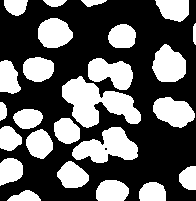} &
   \hspace{-0.3cm}\includegraphics[width=0.125\linewidth]{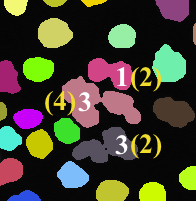} &
   \hspace{-0.3cm}\includegraphics[width=0.125\linewidth]{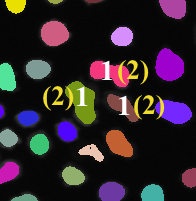} &
   \hspace{-0.3cm}\includegraphics[width=0.125\linewidth]{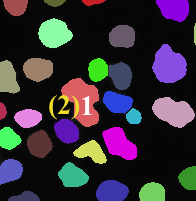} &
   \hspace{-0.3cm}\includegraphics[width=0.125\linewidth]{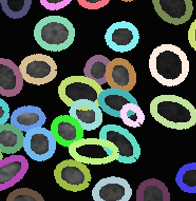} & 
   \hspace{-0.3cm}\includegraphics[width=0.125\linewidth]{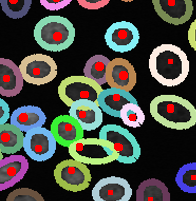} 
\\
   
    \hspace{-0.0cm}\rotatebox{90}{\hspace{0.75cm} t=77}&  
   \hspace{-0.3cm}\includegraphics[width=0.125\linewidth]{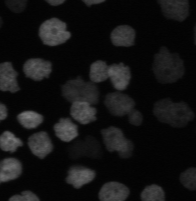} &
   \hspace{-0.3cm}\includegraphics[width=0.125\linewidth]{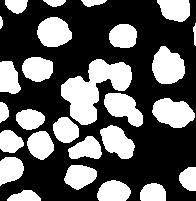} &
   \hspace{-0.3cm}\includegraphics[width=0.125\linewidth]{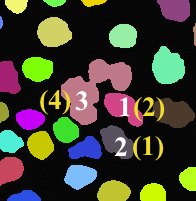} &
      \hspace{-0.3cm}\includegraphics[width=0.125\linewidth]{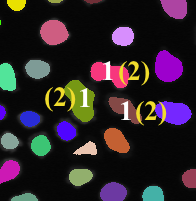} &
   \hspace{-0.3cm}\includegraphics[width=0.125\linewidth]{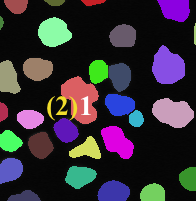} &
   \hspace{-0.3cm}\includegraphics[width=0.125\linewidth]{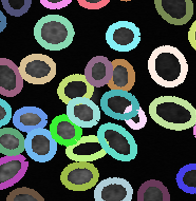} &
   \hspace{-0.3cm}\includegraphics[width=0.125\linewidth]{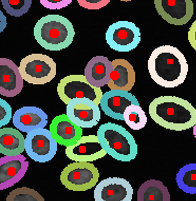} 
\\
   
    \hspace{-0.0cm}\rotatebox{90}{\hspace{0.75cm} t=78}&
   \hspace{-0.3cm}\includegraphics[width=0.125\linewidth]{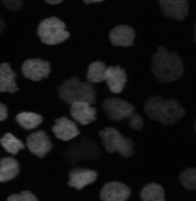} &
   \hspace{-0.3cm}\includegraphics[width=0.125\linewidth]{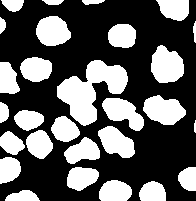} &
   \hspace{-0.3cm}\includegraphics[width=0.125\linewidth]{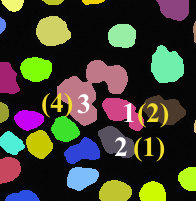} &
      \hspace{-0.3cm}\includegraphics[width=0.125\linewidth]{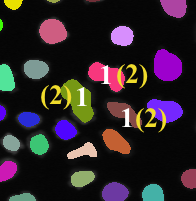} &
   \hspace{-0.3cm}\includegraphics[width=0.125\linewidth]{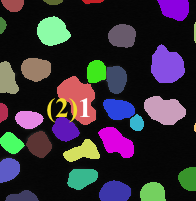} &
   \hspace{-0.3cm}\includegraphics[width=0.125\linewidth]{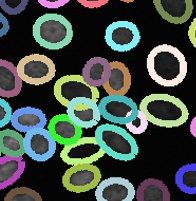} &
   \hspace{-0.3cm}\includegraphics[width=0.125\linewidth]{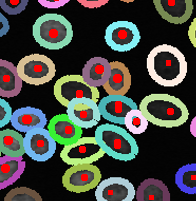} 
\\

\end{tabular}
   \vspace{-6.0mm}
\end{center}
\caption{Three  images  from  a  typical sequence;  the  original  segmentations
  produced  by  a  pixel-based classifier;  the  results
  of~\cite{Schiegg13}(\ct{}),~\cite{Schiegg14}(\jst{}),~\cite{Magnusson12}(\kth{}),    and   our    method
  (\ours{}); the manually annotated tracking ground truth in red dots and the 
  optimal ellipse-tracks obtained using this ground truth. 
  The track identities are encoded in colors. 
  Our approach correctly tracks the cells in spite of long-term segmentation failures,
  and produce results that are very similar to the ground truth. 
  In the \ct{}, \jst{} and \kth{} columns, 
  the white-colored numbers indicate the tracker-inferred numbers of cells that are
  contained by the segments beneath them, and the yellow-colored numbers show 
  the ground truth. Best viewed in color.
  }   
   \vspace{-4.0mm}
   \label{fig:motivation}
\end{figure*}

\comment{
  and the output of several competing methods. ,  Conservation Tracking
     , KTH Cell Tracker (), our
     results (\ours{}) and the ground-truth labels.  For \kth{}, \ours{} and the
     ground truth, the track ID is encoded by colors. For \ct{}, since the final
     tracks    are    not    available     in    the    current    release    of
     Ilastik, we show  the number of inferred cells  in case of
     undersegmentation.  \ct{} in  this case  incorrectly infers  the number  of
     cells in all the three segments,  while \kth{} fails in the segment colored
     orange  in  the  middle.   Our  tracker correctly  tracks  cells  that  are
     undersegmented for consecutive frames. }

%% file: figs/ellipses.tex

\begin{figure}[t!]
\begin{center}
\setlength{\tabcolsep}{1.1pt}
\begin{tabular}{cccc}
\includegraphics[width=0.24\columnwidth]{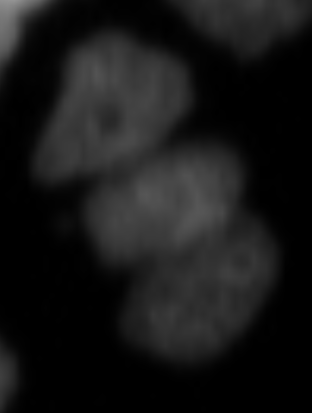} &
\includegraphics[width=0.24\columnwidth]{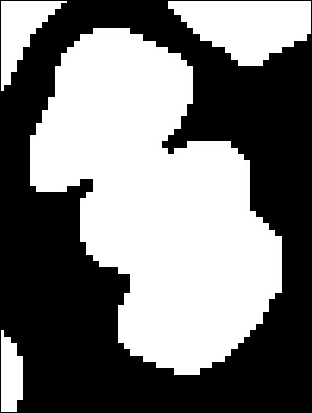} &
\includegraphics[width=0.24\columnwidth]{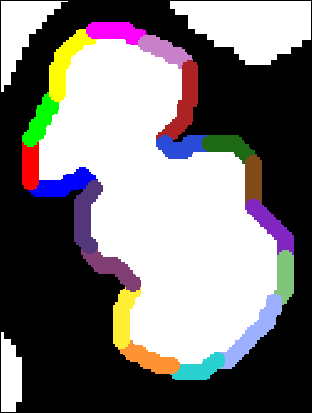} &
\includegraphics[width=0.24\columnwidth]{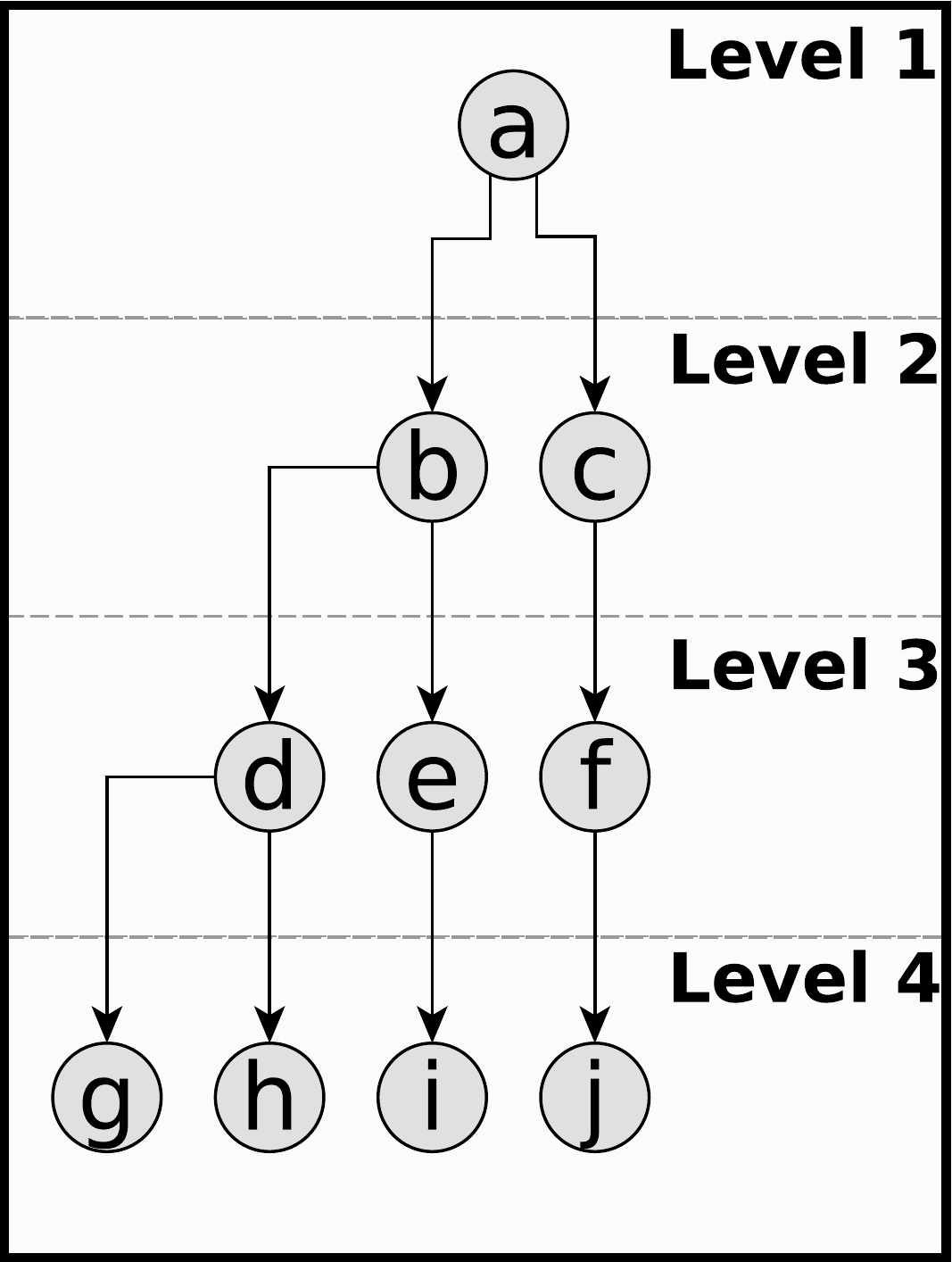} 
\vspace{-0.3em}\\
\small (a) Image & \small (b) Segmentation & \small (c) Contourlets & \small 
(d) Hierarchy
\vspace{0.5em}\\ 
\includegraphics[width=0.24\columnwidth]{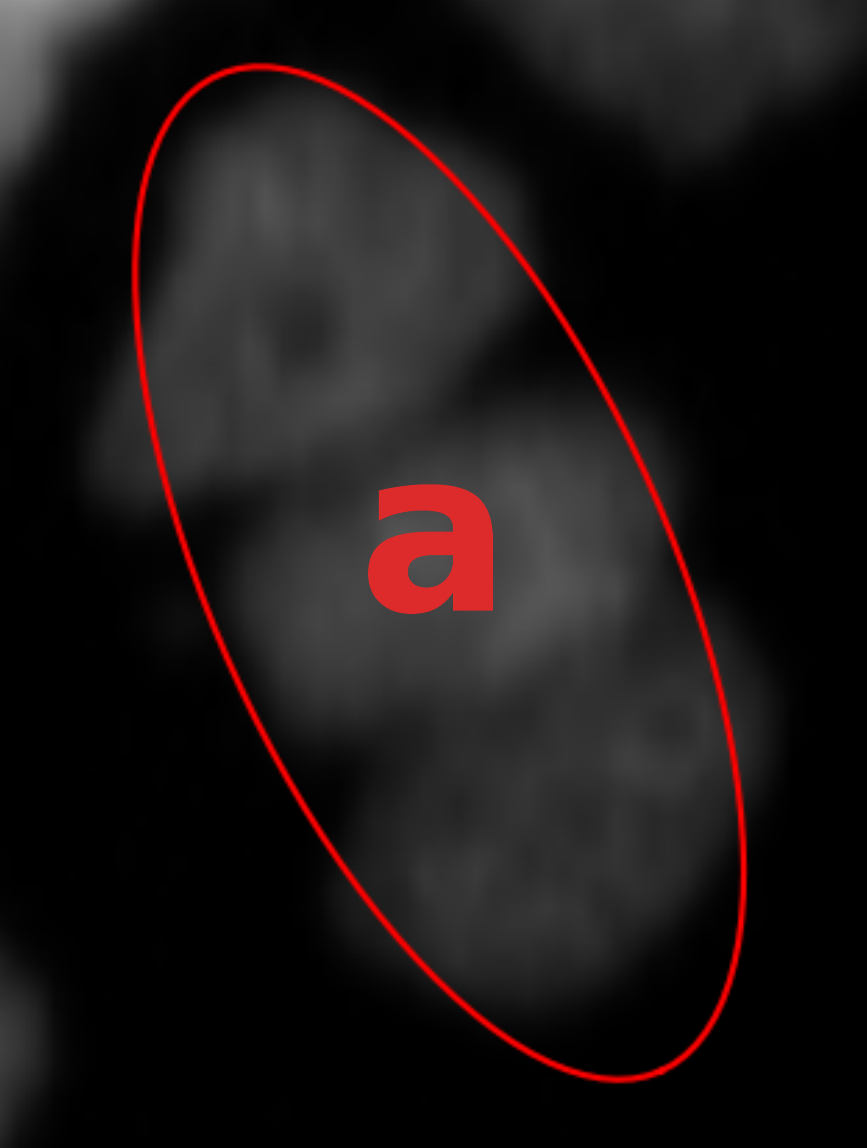} &
\includegraphics[width=0.24\columnwidth]{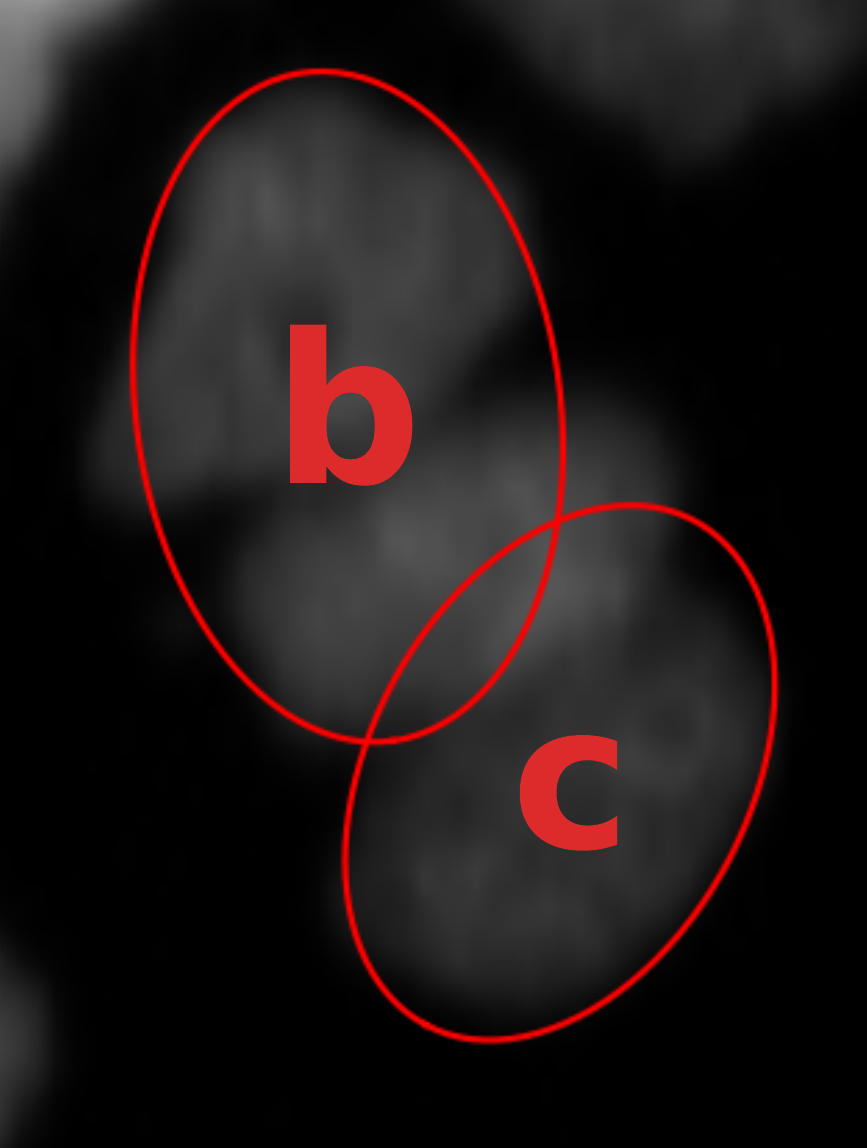} &
\includegraphics[width=0.24\columnwidth]{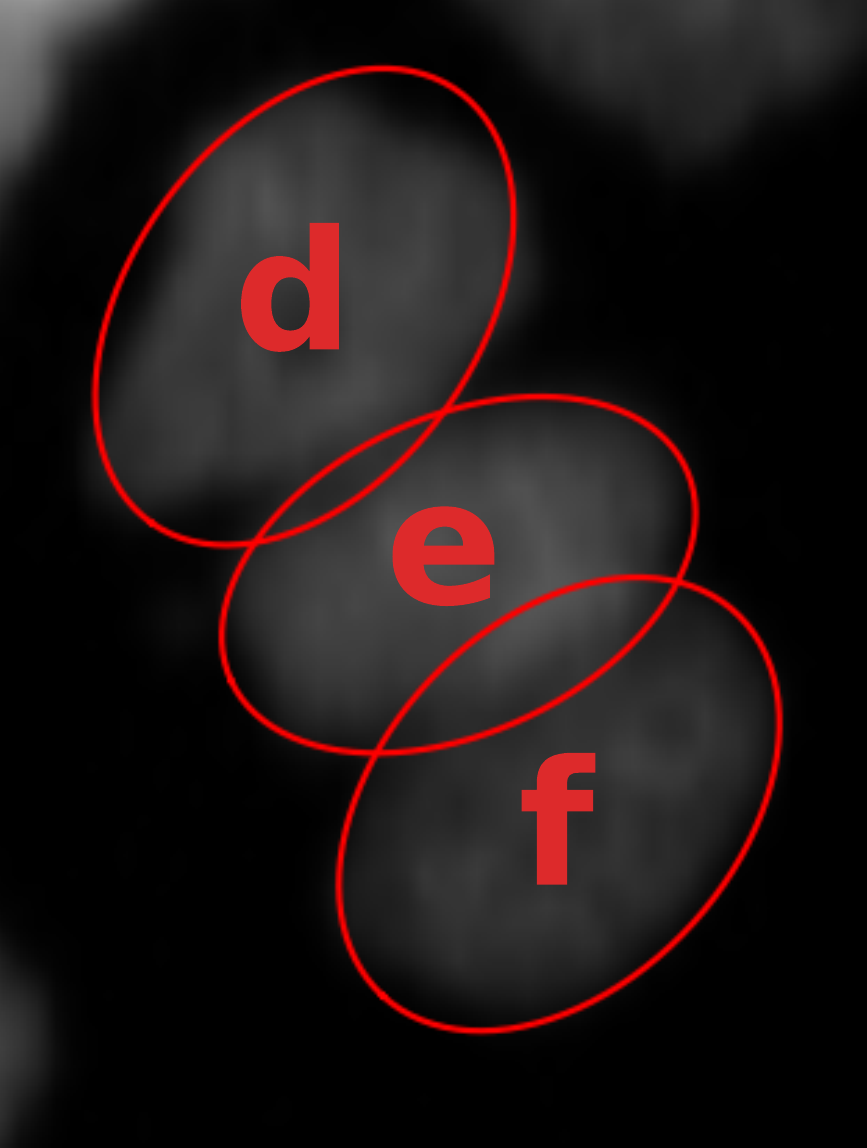} &
\includegraphics[width=0.24\columnwidth]{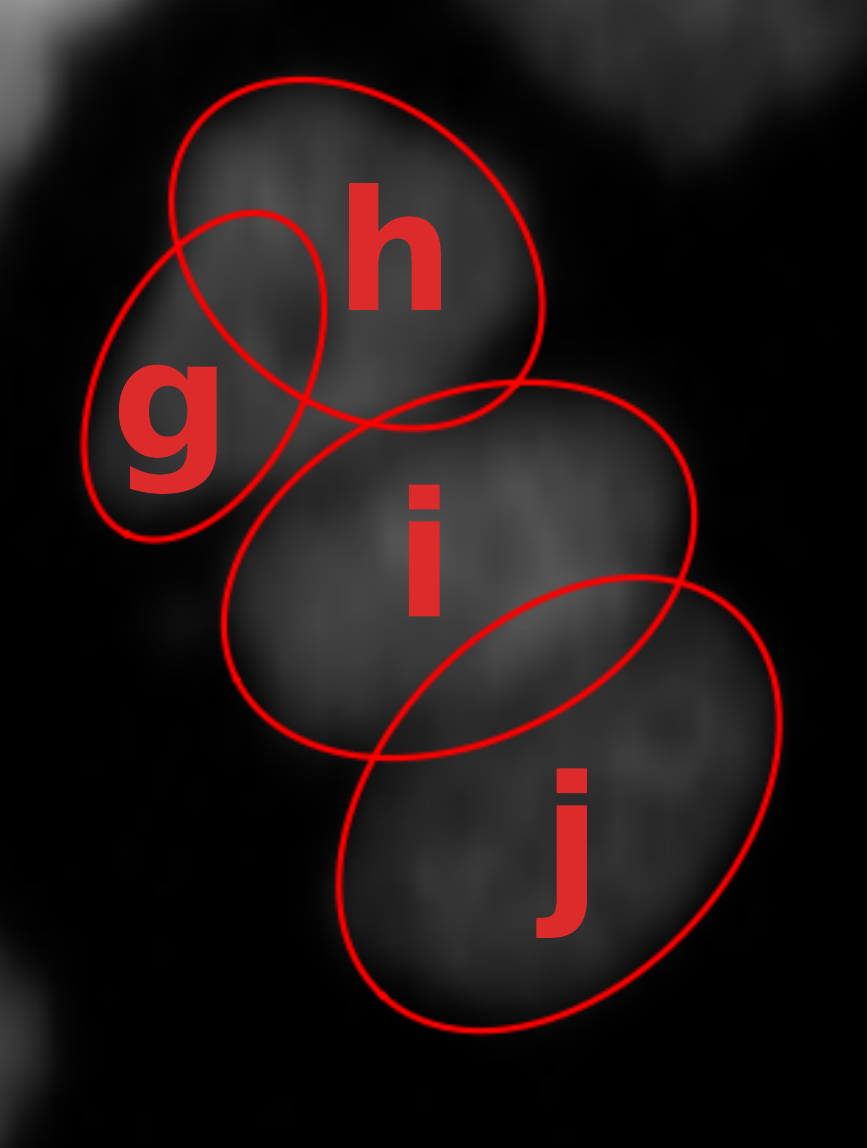}
\vspace{-0.3em}\\
\small (e) Level 1 & \small (f) Level 2 & \small (g) Level 3 &
 \small (h) Level 4\\
\end{tabular}
\end{center}
\vspace{-0.6cm}
\caption{Hierarchy of detection hypotheses. (a) An image region containing three
  HeLa  cells clumped  together.  (b)  Applying a  pixel  classifier results  in
  under-segmentation,  in which  the three  cells appear  as a  single connected
  component.  (c) Automatically extracted  contourlets for this component.  Each
  one is overlaid  in a different color.  They are used to fit  ellipses using a
  hierarchical agglomerative clustering algorithm.   (d) The resulting hierarchy
  of 10  hypotheses. We show only  the first four levels  for simplicity.  (e-h)
  Individual levels  with one, two,  three and  four hypotheses.
}
\vspace{-0.65cm}
\label{fig:close_up_hierarchy}
\end{figure}


%% file: related.tex

\section{Related Work}
\label{sec:related}

Current  tracking  approaches  can   be  divided  into~\emph{Tracking  by  Model
  Evolution}  and~\emph{Tracking   by  Detection}~\cite{Maska14}.    We  briefly
discuss state-of-the-art  representatives of these  two classes below  and refer
the    interested     reader    to    the    much     more    complete    recent
surveys~\cite{Meijering12,Maska14}.

\subsection{Tracking by Model Evolution}

Most algorithms in this class simultaneously track and detect objects greedily 
from frame to frame.   This   means  extrapolating results obtained in earlier 
frames to process the current one, which can be done at  a low  computational  
cost  and is  therefore  fast  in practice.  Such methods have attracted 
attention both in the cell tracking 
field~\cite{Dzyubachyk10,Dufour11,Delgado12,Maska13} as well as in the
more general object tracking one~\cite{Zhang14b,Kalal11,Oron14,Zhang14,Gall11}.
Common techniques in the cell tracking category involve evolving appearance 
or geometry models from one frame to the next, typically done using 
active contours~\cite{Dzyubachyk10,Dufour11,Delgado12,Maska13} or Gaussian 
Mixture Models~\cite{Amat14}.
Though these methods are attractive and mathematically sound, performance 
suffers from the fact  that they only consider  a restricted temporal context 
and therefore cannot  guarantee consistency over a whole  sequence.

This limitation has been addressed by more global active contour 
methods~\cite{Li08,Padfield09} that consider the whole spatio-temporal domain 
to segment the cells and recover parts of their trajectories. Although this provides improved 
robustness at the cost of increased computational burden, these approaches 
do not provide global optimality either.

\comment{In      practice,      only      three      tracking by evolution 
approaches      entered         the   challenges~\cite{Maska14,Solorzano14} and  
were outperformed by  the approaches  discussed below.}

\subsection{Tracking by Detection}

Approaches    in    this    class     have    proved    successful    at    both
people~\cite{Ge08,Collins14,Andriyenko12,Yang12a,Wojek13,Jiang07}    and    cell
tracking~\cite{Li08b,Kausler12,Schiegg13}.

They involve  first detecting the target  objects in individual frames  and then
linking  these  detections to  produce  full  trajectories.  This  is  typically
computationally more  expensive than Tracking  by Model Evolution.   However, it
also tends to  be more robust because trajectories are  computed by minimizing a
global   objective   function   that   enforces   consistency   of   appearance,
disappearance,  and division  over  time.  This can  be  seen  in the  benchmark
of~\cite{Maska14} in which these methods tended to dominate.

One   way  to   perform  tracking-by-detection   is  to   reason  in   the  full
spatio-temporal  grid formed  by  stacking  up all  possible spatial locations  over
time~\cite{Berclaz11,Andriyenko10,Pirsiavash11,Heibel13,Kim12b}.  This can be done 
optimally in polynomial time for  non-dividing objects  such as  pedestrians and  has made  
this approach competitive  despite the large size of  the graphs  involved.  However,  for dividing
objects, the resulting optimization problem is NP-Hard, which is why this dense spatio-temporal approach 
has not been explored for cell tracking.

Instead, practical  tracking-by-detection algorithms for  cells rely on  a small
number  of strong  detections that  can  be later linked  into complete  trajectories.
Recent  efforts  have  focused  on  solving  two  main  challenges  specific  to
cell-tracking, which we discuss below. 

\emph{Division and disappearance.} Cells can  divide or die and disappear. While
rule-based approaches have been used to  handle this, most recent ones formulate
tracking  as a  global  IP~\cite{Kausler12,Schiegg13}.
This  makes  it  possible  to  use   priors  for  cell  migration,  division  and
disappearance  between adjacent  frames.  One  notable exception  is the  method
of~\cite{Magnusson12} that relies on the  Viterbi algorithm to sequentially add
trajectories to  a cell  lineage tree. Motion,  division, and  disappearance are
encoded through  a scoring function  that quantifies  how well the  lineage tree
explains  the  data.   This  algorithm  scored highest  in  one  of  the
  benchmarks~\cite{Maska14}  and we  will  use it  as one  of  our baselines  in
  Section~\ref{sec:exp}.

\emph{Clumped  cells.} There  is  no  guarantee that  individual  cells will  be
detected as separate entities  in any given frame because two  or more cells can
clump together,  producing under-segmentation errors. Many  heuristics have been
proposed  to solve  this  problem.  One  is  to assume  that  clumped cells  are
unlikely to happen  for a specific modality  or that they do not  pose a problem
for the  tracker~\cite{Li08b,Kausler12,Magnusson12}.  While this  is appropriate
in some  cases, it  is clearly invalid  for certain modalities  such as  the one
shown   in   Fig.~\ref{fig:motivation}.   
Other heuristics  involve splitting segmentations using the
Radon and watershed  transforms~\cite{Dzyubachyk10,Maska14}. Unfortunately, they
are still relatively  prone to over- and under-segmentation  that complicate the
tracking task.  
The approach of~\cite{Khan14} generates oversegmentation by fitting ellipses
on Hessian Images. However, this approach focuses on developing embryo and 
relies on strong assumptions, such as assuming that the number of cells do not decrease
from time $t$ to $t+1$.

An ingenious approach is that of~\cite{Schiegg13}, which first  finds trajectories 
by treating segmentations in each  frame as clumps of one or more
cells, with the  exact number being initially unknown, and  then uses a factor
graph  to  resolve this  ambiguity. 
{However, since the final linking is decoupled from the cell cluster tracking, 
the cell count inaccuracies may propagate to the final trajectories.}
\comment{However,  because  the two  steps of 
cell count reasoning and tracking are  decoupled, global optimality is not guaranteed.} 
To overcome this problem,~\cite{Schiegg14} introduces a set of hypotheses, 
  generated by over-segmenting the image into superpixels and 
  subsequently merging them to create competing explanations. 
  Similarly,~\cite{Jug14b} 
  generates the hypotheses using component trees and graph cut.
  However, neither method uses spatially overlapping hypotheses and 
  may result in irrecoverable segmentation errors.
      By contrast, our approach 
  naturally handles such cases, as depicted by Fig.~\ref{fig:close_up_hierarchy}.
  Furthermore, as we will discuss in Section~\ref{sec:NetworkFlow},  our formulation provides a 
  more compact integer program with much less constraints and variables.

%% file: method.tex

\section{Method}
\label{sec:method}

\input{./figs/graph.tex}

\comment{
Our  approach   involves  building   a  spatio-temporal  graph   of  
conflicting  detection hypotheses in individual frames
}

Our  approach   involves  building   a  spatio-temporal  graph   of  
conflicting  detection hypotheses and then finding the 
globally optimal trajectories in it. More specifically, we first produce
a binary image of the underlying  cell populations using a classifier trained on
a few  hand-annotated segmentations.  For  each connected component,  we produce
multiple  hypotheses  by  hierarchically fitting  a varying  number  of
ellipses to it.  This  results in a directed graph, such as  the one depicted by
Fig.~\ref{fig:complete_graph}. Its  nodes are individual ellipses  and its edges
connect  nearby ones  in  consecutive  frames.  Full  trajectories  can then  be
obtained by  solving an integer  program with a  small number of  constraints that
exclude  incompatible  hypotheses  and  enforce consistency  while  allowing  for
cell-division, migration and death.

In the following, we first describe our approach to segmenting cell images and building 
hypotheses graphs from them. We then formulate the simultaneous 
detection and tracking problem on these graphs as a constrained network flow 
problem, and discuss how we compute the various energy terms 
of its objective function.

\subsection{Building Hierarchy Graphs}
\label{sec:Hierarchy}

Our algorithm, like those of~\cite{Li08b,Kausler12,Schiegg13}, starts by segmenting 
cells using local image features. 
To this end, we first train the binary random forest pixel classifier of~\cite{Fiji12} 
for each evaluation dataset on a few partially annotated images.
We use four different types of low-level features: pixel intensities, gradient and hessian 
values, and difference of Gaussians, all of which are computed by first Gaussian 
smoothing the input image with a range of sigma values.

Applying the resulting classifier to the full image sequences results in 
segmentations which often contain groups of clumped cells, such as the ones 
shown in Fig.~\ref{fig:motivation}. Therefore, each connected component of the 
segmentation 
potentially contains an a priori unknown number of cells.

We produce a hierarchy of conflicting detection hypotheses for each such component by fitting a varying number of ellipses to its contours. More specifically, we first identify all the contour points that are local maxima of curvature magnitude. This is done iteratively by selecting the maximum curvature points and suppressing their local neighborhoods. We then break the contour into short segments at these points, which yields a number of contourlets as shown in Fig.~\ref{fig:close_up_hierarchy}(c). We cluster them in a hierarchical agglomerative fashion and fit ellipses to each resulting cluster using the non-iterative least squares approach of~\cite{Prasad13}. In all our experiments, we set the size of the suppression neighborhood to seven pixels because this is the minimum number of points required to reliably fit an ellipse using this approach.

Let $C$ denote the set of all contourlet clusters for a connected component. Given a pair of clusters $C_i \in C$ and $C_j \in C$, we define their distance to be%

\vspace{-6mm}
\begin{small}
\begin{equation}
\bigg[
\sum_{C_l \in \{ C_i \cup C_j \}}{\hspace{-15pt} h(C_l,e)} +
\hspace{-10pt}  \sum_{C_l \in C \setminus \{ C_i \cup C_j \}}{\hspace{-15pt} g(C_l,e)}
\bigg] 
c(e)
\sqrt{\frac{1}{1+ec^2(e)}} \ ,
\label{eq:cluster_dist}
\end{equation}
\end{small}%
\noindent where $e$ is the ellipse obtained by fitting to the points of $C_i \cup C_j$, and $c(e)$ and $ec(e)$ are its circumference and eccentricity respectively. $h(C_l,e)$ denotes the Hausdorff distance between the points of $C_l$ and the ellipse $e$. The function $g(C_l,e)$ is defined in a similar way but without considering the points of $C_l$ that are outside $e$.

The first term in the product captures image evidence along the contours of the entire connected component while the last two ones act as a shape regularizer to prevent implausible ellipse geometries from appearing in the solution. We use this distance measure to compute an ellipse hierarchy, such as the one of Fig.~\ref{fig:close_up_hierarchy}, for every connected component in the  temporal sequence.

Given these over-complete hierarchies, we then build a graph, whose 
vertices are the ellipses and the edges link pairs of them that belong to two 
spatially close connected components in consecutive frames.
The resulting graph has a hierarchical dimension that allows for finding the globally 
optimal cell detections and trajectories in a single shot.

\comment{This is similar in spirit to recent graph-based 
approaches~\cite{Kausler12,Schiegg13}, but with the key difference that our 
graphs have a hierarchical dimension that allows for finding the globally 
optimal detection hypotheses and trajectories in a single shot.
}

\subsection{Network Flow Formalism}
 \label{sec:NetworkFlow}
 
The procedure described above yields a directed graph $G'=(V',E')$, which we then augment 
with three distinguished vertices; namely the source $s$, the sink $t$ and the division 
$d$ as depicted by Fig.~\ref{fig:complete_graph}. We connect these three vertices to 
every other vertex in $G'$ to allow accounting for cell appearance, disappearance and division 
 in mid-sequence. 

Let $G=(V,E)$ be the resulting graph obtained after the augmentation.  We define a binary flow variable $f_{ij}$ for each edge $e_{ij} \in E$ to indicate the presence of any one of the following cellular events:

\vspace{-1mm}
\begin{packed_item}
 \item  Cell migration from vertex $i$ to vertex $j$,
  \item  Appearance at vertex $j$, if $i = s$,
  \item  Division at vertex $j$, if $i = d$,
  \item  Disappearance at vertex $i$, if $j = t$.
\end{packed_item}

\vspace{-1mm}
Let $\mathbf{f}$ be the set of all $f_{ij}$ flow variables and $\mathbf{F} = \{ F_{ij} \}$ be the set of all corresponding hidden variables.
Given an image sequence $\mathbf{I}\!= (\mathbf{I}^1, \dots, \mathbf{I}^T)$ with $T$ temporal frames and the corresponding graph $G$,  we look for the optimal trajectories $\mathbf{f}^*$ in $G$ as the solution of
\begin{small}
\begin{eqnarray}
\hspace{-3.0mm} \mathbf{f}^*  & \hspace{-2mm} =  &  \hspace{-2mm} \underset{\mathbf{f} \in {\cal F}}{\operatorname{argmax}}  \,  P(\mathbf{F} = \mathbf{f} \mid \, \mathbf{I})  \label{eq:TruePosterior} \\
 & \hspace{-2mm} \approx & \hspace{-2mm} \underset{\mathbf{f} \in {\cal F}}{\operatorname{argmax}} \prod_{e_{ij} \in E} P(F_{ij} = f_{ij} \mid \mathbf{I}^{t(i)}, \mathbf{I}^{t(j)})  \label{eq:DiscreteOpt2} \\
& \hspace{-2mm} = &  \hspace{-2mm} \underset{\mathbf{f} \in {\cal F}}{\operatorname{argmax}} \prod_{e_{ij} \in E} P(F_{ij} = 1 \mid \mathbf{I}^{t(i)}, \mathbf{I}^{t(j)})^{f_{ij}}  \times \nonumber \\ 
& \hspace{-2mm} & \hspace{49pt}  P(F_{ij} = 0 \mid \mathbf{I}^{t(i)}, \mathbf{I}^{t(j)})^{(1 - f_{ij})}  \label{eq:DiscreteOpt3} \\
& \hspace{-2mm} = & \hspace{-2mm} \underset{\mathbf{f} \in {\cal F}}{\operatorname{argmax}} \sum_{e_{ij} \in E} \log \hspace{-0.5mm} \left( \frac{P(F_{ij} = 1 \mid \mathbf{I}^{t(i)}, \mathbf{I}^{t(j)})}{P(F_{ij} = 0 \mid \mathbf{I}^{t(i)}, \mathbf{I}^{t(j)})} \right) \hspace{-0.5mm} f_{ij} \,        \label{eq:DiscreteOpt4} \\
& \hspace{-2mm} = & \hspace{-2mm} \underset{\mathbf{f} \in {\cal F}}{\operatorname{argmax}} \hspace{-0.5mm} \sum_{e_{ij} \in E'} \hspace{-1mm} \log \hspace{-0.5mm} \left( \hspace{-0.75mm} \frac{\rho_{ij}}{1\hspace{-0.5mm} - \hspace{-0.5mm} \rho_{ij}} \hspace{-0.75mm} \right) \hspace{-0.5mm} f_{ij} \hspace{-0.5mm}  +  \sum_{j \in V} \log \hspace{-0.5mm} \left( \hspace{-0.75mm} \frac{\rho_{a}}{1 \hspace{-0.5mm} - \hspace{-0.5mm} \rho_{a}} \hspace{-0.75mm} \right) \hspace{-0.5mm} f_{sj} \hspace{-0.5mm}  \nonumber \\
& \hspace{-2mm} & +  \sum_{i \in V} \log \hspace{-0.5mm} \left( \hspace{-0.75mm} \frac{\rho_{d}}{1\hspace{-0.5mm} - \hspace{-0.5mm} \rho_{d}} \hspace{-0.75mm} \right) f_{it} + 
\sum_{j \in V} \log \hspace{-0.75mm} \left( \hspace{-0.5mm} \frac{\rho_{j}}{1 \hspace{-0.5mm} - \hspace{-0.5mm} \rho_{j}} \hspace{-0.5mm}  \right) \hspace{-0.5mm} f_{dj} \label{eq:DiscreteOpt5} \ ,
\end{eqnarray}
\end{small}%
where $\mathbf{I}^{t(i)}$ is the temporal frame containing vertex $i$, and ${\cal F}$ denotes the set of all feasible cell trajectories, which satisfy linear constraints. In Eq.~\ref{eq:DiscreteOpt2}, we assume that the flow variables $F_{ij}$ are conditionally independent given the evidence from consecutive frame pairs.
Eqs.~\ref{eq:DiscreteOpt3} and~\ref{eq:DiscreteOpt4} are obtained by using the 
fact that the flow variables are binary and by taking the logarithm of the 
product. Finally, in Eq.~\ref{eq:DiscreteOpt5}, we split the sum into four parts 
corresponding to the four events mentioned above.

The appearance and disappearance probabilities, $\rho_{a}$  and $\rho_{d}$, are computed simply by finding the relative frequency of these events in the ground truth cell lineages of the training sequences. On the other hand, the migration and the division probabilities, $\rho_{ij}$ and $\rho_{j}$, are obtained using a classification approach as described in the next section.

We define three sets of linear constraints to model cell behavior and exclude conflicting detection hypotheses from the solution, which we describe in the following.
\paragraph{Conservation of Flow:} 
We require the sum of the flows incoming to a vertex  to be equal to the sum of the outgoing flows.
This allows for all the four cellular events while incurring their respective costs given in Eq.~\ref{eq:DiscreteOpt5}.
\begin{small}
\begin{eqnarray}
	\sum_{e_{ij} \in E'}  f_{ij} +  f_{sj} +  f_{dj} = \sum_{e_{jk} \in E'} f_{jk} + f_{jt} \ ,  \hspace{0.5cm} \forall j \in V'  \ .
\end{eqnarray}
\end{small}
\vspace{-0.7cm}
\paragraph{Prerequisite for Division:} 
We allow division to take place at a vertex $j \in V'$ only if there is a cell at that location. We write this as
\begin{small}
\begin{eqnarray}
	\sum_{e_{ij} \in E'}  f_{ij} +  f_{sj} \geq  f_{dj} \ , \hspace{0.5cm} \forall j \in V' \ .
\end{eqnarray}
\end{small}
\vspace{-0.7cm}
\paragraph{Exclusion of Conflicting Hypotheses:}
Given a hierarchy tree of detections, we define an exclusion set $S_l$ for each terminal vertex $l \in V'$ of this tree. For instance, in the example of Fig.~\ref{fig:close_up_hierarchy}(d), the four exclusion sets are $\{a,b,d,g\}$, $\{a,b,d,h\}$, $\{a,b,e,i\}$ and $\{a,c,f,j\}$.
Let $S$ be the collection of all such sets for all the connected components in the sequence. We disallow more than one vertex from each set to appear in the solution, and express this as
\begin{small}
\begin{eqnarray}
	\sum_{\substack{j \in S_l, \\ e_{ij} \in E'}} f_{ij} + \sum_{j \in S_l }  f_{sj} \leq 1, \hspace{0.5cm} \forall S_l \in S \ .
\label{eq:exclusion_constraints}
\end{eqnarray}
\end{small}
\noindent We solve the resulting integer programs within an optimality tolerance 
of $1e^{-3}$ using the branch-and-cut algorithm implemented in the Gurobi 
optimization library~\cite{Gurobi}. 

\comment{indicating that our results are
very close to the global optimal solution.} 

Note that our integer program contains only a single 
  set of variables and three sets of linear constraints, and therefore it is 
more compact than those of the recent approaches of~\cite{Schiegg14} and~\cite{Schiegg13} that are similar  to ours.
Formally, let $N$, $C$, and $K$  denote the total number of detections, 
  total number of exclusion sets and average number of neighbors per detection.
  The total number of constraints for ours and \cite{Schiegg14} are 
  $C+2N$ and $C+2N+2NK$; the total number of variables
  are $N(3+K)$ and $N(1+ K + 2^{(K+1)})$ respectively.
The term $2^{(K+1)}$ comes from the fact that~\cite{Schiegg14} 
  includes indicator variables for higher order factors.

\subsection{Cell Migration and Division Classifiers}
 \label{sec:Classifiers}
 
Given two adjacent vertices $i,j \in V'$, and their associated ellipses 
$e_i$ and $e_j$ in consecutive time frames, we train a 
classifier to estimate the likelihood that both belong to the same cell. 
More specifically, we use a Gradient Boosted Tree (GBT) 
classifier~\cite{Friedman02} to learn a function 
$\varphi_\textrm{migr}(e_i,e_j) \in \mathbb{R}$, 
based on both appearance and geometry features including
the distance between the ellipses, their eccentricities and degree of overlap,   
hierarchical fitting errors and ray features.
Once $\varphi_\textrm{migr}(e_i,e_j)$ is learned, we apply Platt 
scaling to compute the probability $\rho_{ij}$ 
that the two ellipses belong to the same cell, and plug it into Eq.~\ref{eq:DiscreteOpt5}.

Similarly, the likelihood of ellipse $e_j$ at time $t$ dividing into two ellipses $e_{k}$ 
and $e_{l}$ at $t+1$ is learned with another GBT classifier, trained on 
features such as the orientation and size differences among the ellipses. 
We present a detailed list of both the migration and division features in the 
Supplementary Material.

For prediction, we compute the division score for ellipse $e_j$ at time $t$ 
as
\vspace{-1mm}
\begin{equation}
\varphi_\textrm{div}(e_j) = \max_{\substack{e_{jk} \in E', \ e_{jl} \in E': \\ k \neq l}} 
\ \ \ \varphi_\textrm{div}(e_j,e_k,e_l) \:,
\end{equation}%
\noindent where $\varphi_\textrm{div}(\cdot,\cdot,\cdot)$ is the scoring function learned 
by the classifier, and $(e_k, e_l)$ is a pair of ellipses corresponding to two 
potential daughter cells at time $t+1$. We obtain the division probability $\rho_{j}$  
of Eq.~\ref{eq:DiscreteOpt5} from $\varphi_\textrm{div}(e_j)$, again using Platt scaling.

%% file: figs/graph.tex

\newcolumntype{V}{>{\centering\arraybackslash} m{0.5\linewidth} }
\begin{figure*}[t]
\begin{center}
\begin{tabular}{VV}
   \includegraphics[width=0.62\linewidth]{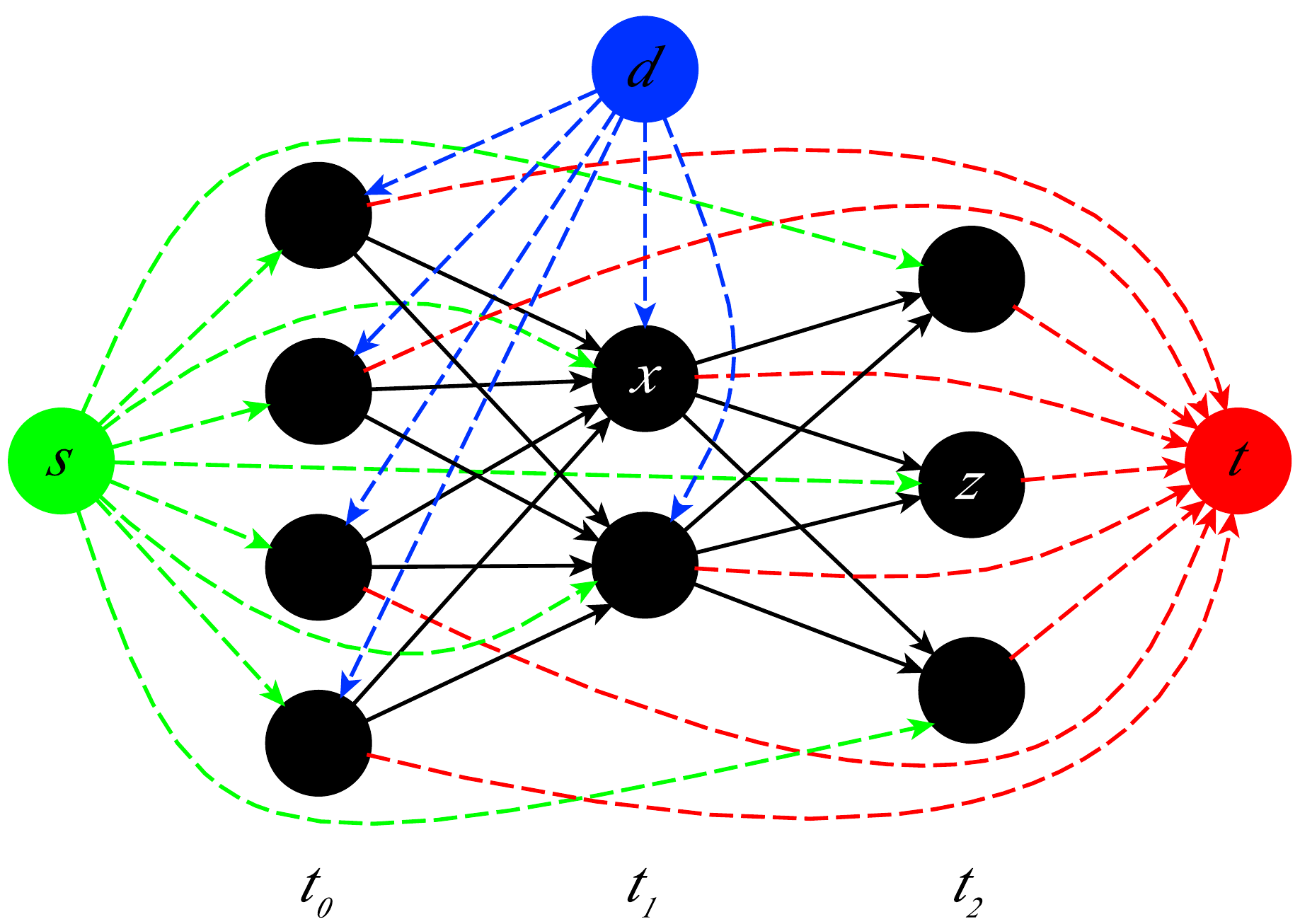} &
   \includegraphics[width=0.62\linewidth]{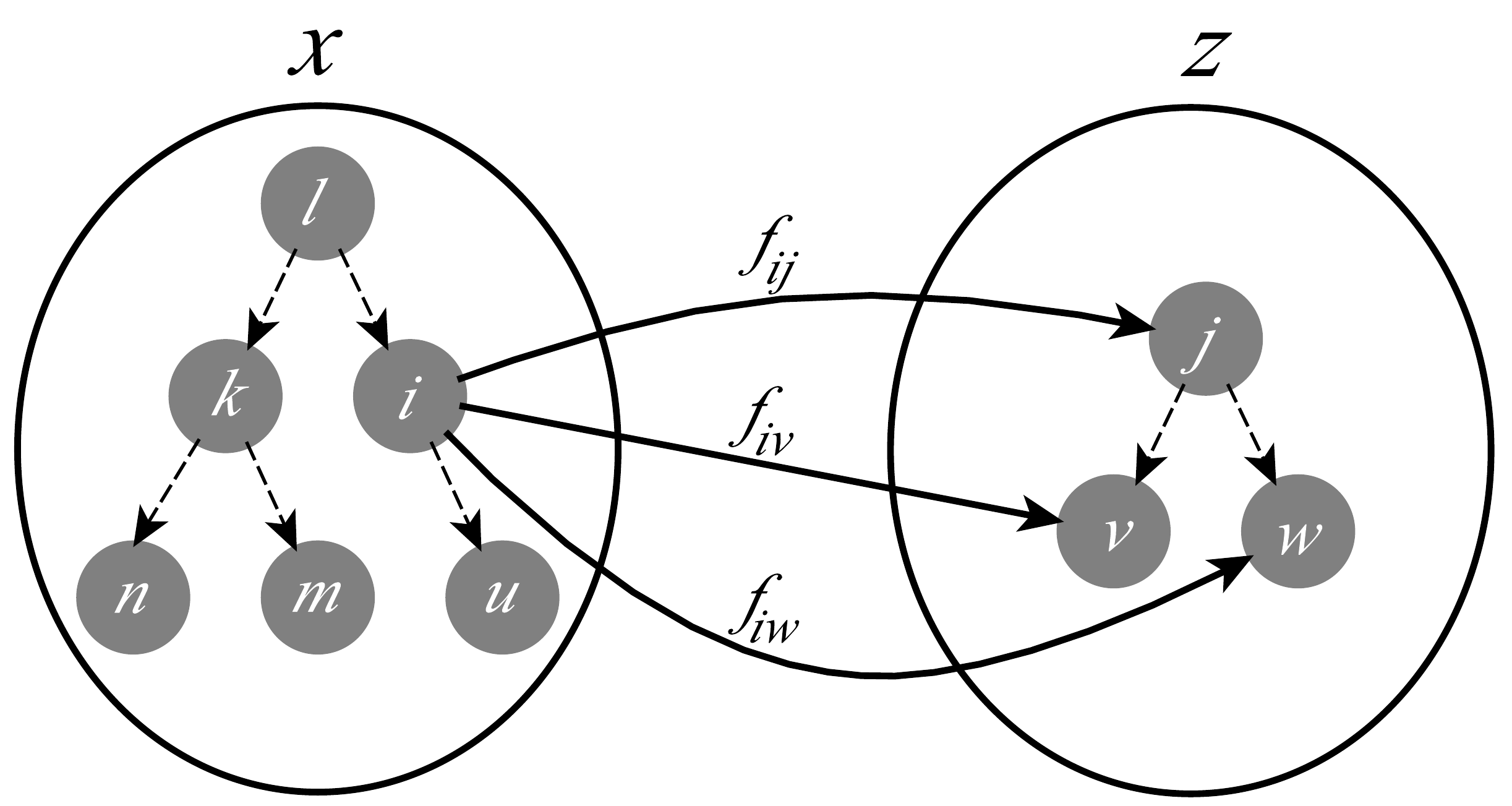} \\
    (a) & (b) 
\end{tabular}
   \vspace{-7.0mm}
\end{center}
   \caption{Spatio-temporal graph  of hypotheses for 3  consecutive time frames.
     (a)  Each  black circle  is  a  hypervertex  corresponding to  a  connected
     component of the  segmentation and is connected to neighboring  ones at the
     next time step.  The special vertices $s$ (source, in green), $t$ (sink, in
     red) and  $d$ (division, in  blue) allow respectively for  cell appearance,
     disappearance and  division.  (b)  Each hypervertex, such  as $x$  and $z$,
     contains a hierarchical set of hypotheses (vertices) such as those depicted
     by Fig.~\ref{fig:close_up_hierarchy}.   These hypotheses are shown  as gray
     circles and  connected to nearby ones  in the following frame  via directed
     edges.   We only  show three  of  these edges  to avoid  clutter.  On  each
     hypervertex, we define an exclusion set for each leaf vertex.  For example,
     we define three exclusion sets on $x$: $S_n=\{l,k,n\}$, $S_m=\{l,k,m\}$ and
     $S_u=\{l,i,u\}$. We allow  the tracker to select at most  one vertex within
     each one.  Best viewed in color. }
   \label{fig:complete_graph}
   \vspace{-5.0mm}
\end{figure*}

%% file: exp.tex

\section{Experiments}
\label{sec:exp}

In this section,  we first introduce the datasets  and state-of-the-art 
methods we use as baselines for evaluation purposes.  We then demonstrate  that our approach
significantly outperforms these  baselines, especially when the  cells divide or
are  not well  separated  in the  initial segmentations.   Our  software can  be
downloaded from {\it an  URL to be specified in the final  version of the paper}
and the corresponding tracking videos are available as supplementary material.

\subsection{Test Sequences}
\label{sec:testSeq}

We  used  10   image  sequences  from  three  datasets  of   the  cell  tracking
challenge~\cite{Solorzano14}.  They involve multiple cells that migrate, appear,
disappear, and divide. Difficulties arise from low contrast to the background,
complex  cell  morphology,   and  significant  mutual  overlap.    We  used  the
leave-one-out  training and  testing scheme  within  each dataset  to train  the
classifiers of Section~\ref{sec:Classifiers} and to learn the appearance
and disappearance probabilities of Eq.~\ref{eq:DiscreteOpt5}.

\begin{itemize}
  
\item{\bf HeLa Dataset:} It comprises  two 92-frame sequences from the MitoCheck
  consortium. Cell  divisions are  frequent, which  produces a  dense population
  with  severe  occlusions.   \comment{The   speckles  of  the  microscope  make
    segmentation and tracking even more challenging. }

 \item{\bf  SIM Dataset:}  It comprises  six 50-  to 100-frame  sequences.  They
   simulate migrating and dividing nuclei on a flat surface.

 \item{\bf  GOWT Dataset:}  It comprises  two 92-frame  sequences of  mouse stem
   cells. Their  appearance varies widely and  some have low contrast  against a
   noisy background.

\comment{Compared to the other two
datasets, there are few division events occurred in this datasets, which makes
it challenging to learn the division classifier. }

\end{itemize}

\input{figs/results.tex}

\input{baselines.tex}

\subsection{Evaluation Metrics}

We use  precision, recall  and the  F-Measure, defined as  the harmonic  mean of
precision  and  recall, to  quantify  the  algorithms'  ability to  detect  cell
division, detection, and migration events. 
We also use two global metrics, multiple object tracking accuracy (MOTA)~\cite{Kasturi09} 
and tracking precision (TRA)~\cite{Solorzano14}, to evaluate the overall tracking performance.

We follow the same evaluation methodology as in~\cite{Schiegg13}, which uses the
connected  components of  the  initial segmentations  to  compute the  division,
detection, and migration  accuracies.  We consider a cell migration  event to be
successfully detected if both connected components  of the cell at $t$ and $t+1$
are  correctly  identified.  Similarly,  we  consider  a  division event  to  be
successfully  detected  if it  occurs  at  the  correct  time instant  with  the
connected components of the parent and both daughter cells correctly determined.
Finally, a detection event is said to be successfully identified if an algorithm
infers the right number of cells within a connected component.

Unlike the  above measures, MOTA and  TRA are defined on  individual cell tracks
rather  than  connected components  that  potentially  contain multiple  clumped
cells. They  therefore provide  a global picture  of tracking  performance.  The
main difference  between TRA and  MOTA is,  TRA defines different  penalties for
different types of errors, while MOTA treat all errors equally.

\input{figs/results_ours.tex}

\subsection{Comparing against the Baselines}

We ran our algorithm and the baselines discussed above on all the test sequences
introduced in Section~\ref{sec:testSeq}.  Table~\ref{tab:results} summarizes the
results  for a  representative subset  and  the remainder  can be  found in  the
supplementary material.

Some  numbers  are  missing  because the  publicly-available  implementation  of
\ct{}~\cite{Schiegg13} we use does  not provide the  identities of individual  cells in
under-segmentation  cases.   We therefore  cannot  extract  the complete  tracks
required    to    compute    the    MOTA   and TRA scores. 
For our method and that of~\ct{}, we computed the detection scores by using 
the same segmentations obtained from the pixel classification approach of Section~\ref{sec:Hierarchy}.
By contrast, \gmm{}~\cite{Amat14}, \kth{}~\cite{Magnusson12} and \jst{}~\cite{Schiegg14} 
trackers take only raw images  as  input and do not accept external segmentations to be used. 
Therefore, we  cannot compute  their  accuracy  
for  the detection events. Finally,  in some cases, \gmm{} generates
non-consecutive cell tracks, which is not accepted by the TRA evaluation software.

Table~\ref{tab:results} shows that our tracker consistently yields a significant
improvement on the division and detection  events.  Even on the migration events
for  which the  baselines  already perform  very well,  we  do slightly  better.
However,  because  the  division  and  detection events  are  rare  compared  to
migrations, the  significant improvements on  these  two events only  have a
small impact on MOTA and TRA.

The comparatively  poor performance of \gmm{} can be  partially ascribed to
the fact  that it  relies on  a simple  hand-designed appearance  and geometry
model to detect individual cells.  \kth{}  relies on a richer appearance model
that  improves  performance  but  requires tuning  more  parameters  for  each
sequence.  \ct{} employs several cell-event classifiers and
solves the tracking problem using integer programming, 
resulting in an overall higher division accuracy.
Finally,  the performance  of \jst{} is  negatively impacted  
by the  fact that it  depends on  a watershed-based
oversegmentation   that is sensitive to inaccuracies in pixel probability estimates. 
By contrast, our  ellipse-fitting approach to generating competing
hypotheses is robust to the ambiguous image evidence. 
Given the same initial segmentation as \ct{}, 
our method achieves the best overall performance,
thanks to the simultaneous detection and tracking.

Our tracker runs relatively fast even though we use a large number of variables.
We show  in Tab.~\ref{tab:results} the  running time  of all trackers  given the
detections.   In the  SIM-4 case,  the number  of flow  variables is  around 1.1
million but the  integer programming optimization takes only 2  seconds.  In the
HeLa-2 case,  the number of variables  is around 8 million  and the optimization
takes about 232 seconds.

\subsection{Evaluating Individual Components}

To produce the  results summarized by Table~\ref{tab:results}, we  used our full
approach      as      described       in      Section~\ref{sec:method}.       In
Table~\ref{tab:results_ours}, we show  what happens when we turn  off some of
its components to gauge their respective impacts.

\cl{} relies on local classifier scores and does not impose temporal consistency. 
That is why, it produces  a large number 
of spurious cell tracks. 
\nc{}  addresses this by imposing temporal consistency but 
it allows multiple conflicting hypotheses to be active simultaneously. Therefore, it still suffers 
from spurious detections, which leads to low precision.
\fd{}  disallows conflicting detections but relies on a fixed division probability, which is why 
it gives low division performance.
\bh{}  uses division classifier costs but collapses the hierarchical dimension of our graphs and results in mis-detections.
Finally, \lp \ removes the integrality constraints on the flow variables. 
This gives a similar performance to \ours \ on most of the sequences suggesting that 
the integrality constraints are seldom helpful. However, in the  case of HeLa-2,  
where division events are frequent,  we  observed  that around  3\% of  the 
non-zero flow variables are  fractional, which explains the 2\% drop in recall 
compared  to \ours.

%% file: figs/results.tex

\begin{table*}[t]
\centering 
\begin{small}
\begin{tabular}{l  l  l || c  c  c | c  c  c | c  c  c | c | c | c} 
 & & & \multicolumn{3}{c}{\bf Division} \vline
& \multicolumn{3}{c}{\bf Detection} \vline
& \multicolumn{3}{c}{\bf Migration} \vline
& \multirow{2}{3em}{\bf MOTA} 
& \multirow{2}{2em}{\bf {TRA}} 
& \multirow{2}{2em}{\bf Time}
\\
 & &  & Rec & Pre. & F-M. & Rec. & Pre. & F-M. & Rec. & Pre. & F-M. &  & \\
\hline
\hspace{-1.8em}\rotatebox{90}{\hspace{-3.7em} \multirow{5}{4.5em}{\bf HeLa-1}} 
&\multirow{5}{6.2em}{\includegraphics[width=0.95\linewidth]{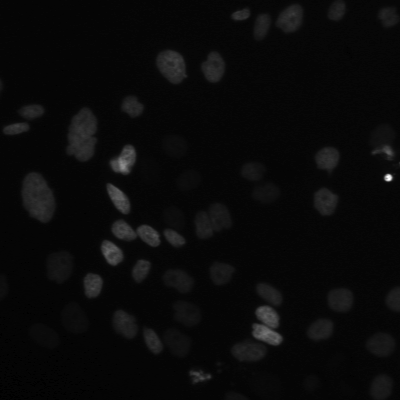}}
& \hspace{-1em} \gmm         & 0.56 & 0.43 & 0.48 & N/A & N/A  & N/A  & 0.92 & 0.98 & 0.95 & 0.82 & {N/A} & {\bf 44}\\
& & \hspace{-1em} \kth         & 0.65 & 0.72 & 0.68 & N/A	& N/A  & N/A  & 0.95 & {\bf 0.99} & 0.97 & 0.91 & \bf{{0.98}}& 70\\
& & \hspace{-1em} \ct 			& 0.74 & {\bf 0.79} & 0.77 & 0.56 & {\bf 0.89} & 0.69 & 0.94 & {\bf 0.99} & 0.97 & N/A & {N/A} & 74.85\\
& & \hspace{-1em} \jst 			& 0.79 & 0.55 & 0.65 & N/A	& N/A  & N/A & 0.86 &  0.92 & 0.89 & 0.73  & 0.80 & 128.16 \\
& & \hspace{-1em} \ours 		& {\bf 0.92} & {\bf 0.79} & {\bf 0.85} & {\bf 0.96} & 0.83 & {\bf 0.89} & {\bf 0.97} & {\bf 0.99} & {\bf 0.98} & {\bf 0.94} &  \bf{{0.98}} & 88.25\\
\hline

\hspace{-1.8em}\rotatebox{90}{\hspace{-3.7em} \multirow{5}{4.5em}{\bf HeLa-2}} 
&\multirow{5}{6.2em}{\includegraphics[width=0.95\linewidth]{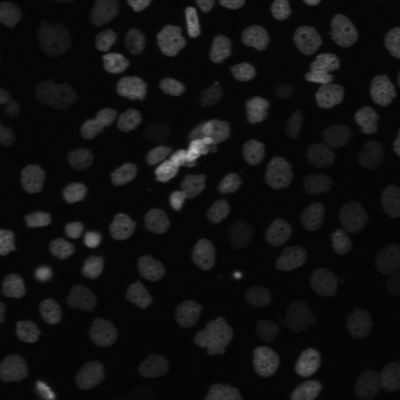}}
 & \hspace{-1em} \gmm         & 0.40 & 0.18 & 0.24 & N/A	& N/A  & N/A  & 0.95 & 0.98 & {\bf 0.97} & 0.43 & {N/A} & {\bf 74}\\
& & \hspace{-1em} \kth 		& 0.65 & 0.72 & 0.68 & N/A  & N/A  & N/A  & 0.94 & {\bf 0.99} & {\bf 0.97} & {\bf 0.90} & \bf{{0.97}}& 336\\
& & \hspace{-1em} \ct 			& 0.76 & 0.81 & 0.78 & 0.73 & 0.63 & 0.67 & 0.94 & {\bf 0.99} & 0.96 & N/A & {N/A} & 79.33 \\
& & \hspace{-1em} \jst 			& 0.69 & 0.44 & 0.54  & N/A	& N/A  & N/A & 0.91 & 0.98 & 0.94 & 0.82 & 0.85 & 88.79\\
& & \hspace{-1em} \ours 		& {\bf 0.86} & {\bf 0.83} & {\bf 0.84} & {\bf 0.86} & {\bf 0.78} & {\bf 0.82} & {\bf 0.96} & {\bf 0.99} & {\bf 0.97} & {\bf 0.90} & \bf{{0.97}} &  232.31 \\
\hline

\hspace{-1.8em}\rotatebox{90}{\hspace{-4.3em} \multirow{5}{5.1em}{\bf GOWT-2}} 
&\multirow{5}{6.2em}{\includegraphics[width=0.95\linewidth]{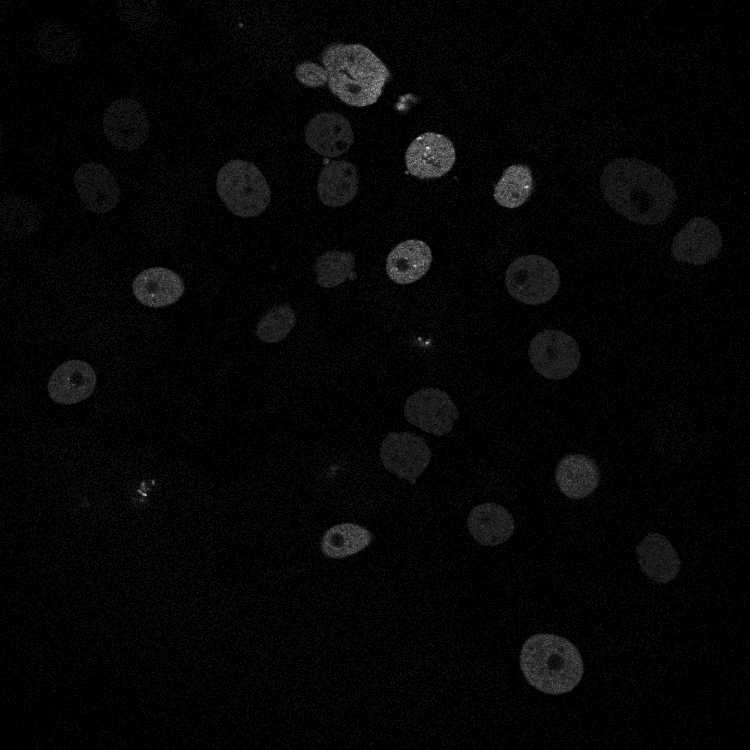}}
 & \hspace{-1em} \gmm              & 0.0 & 0.0 & 0.0 & N/A	& N/A  & N/A  & 0.16 & 0.79 & 0.26 & 0.02 & {N/A} & 37\\
& & \hspace{-1em} \kth 			& 0.0  & N/A  & 0.0  & N/A  & N/A  & N/A  & 0.94 & {\bf 1.0} & 0.97 & 0.94 &  {{0.91}} & 16\\
& & \hspace{-1em} \ct 			& {\bf 1.0} & 0.17 & 0.29 & {\bf 1.0}  & 0.02 & 0.03 & 0.95 & {\bf 1.0} & 0.97 & N/A & {N/A} & 0.81\\
& & \hspace{-1em} \jst 			& {\bf 1.0} & 0.02  & 0.04  & N/A	& N/A  & N/A & 0.93 & 0.98  & 0.95 & 0.85 & {\bf 0.95} & 3.87\\
& & \hspace{-1em} \ours 		& {\bf 1.0}  & {\bf 1.0}  & {\bf 1.0}	 & {\bf 1.0}  & {\bf 1.0}  & {\bf 1.0}  & {\bf 0.95} & {\bf 1.0} & {\bf 0.98} & {\bf 0.96} &\bf{{0.95}} & {\bf 0.22}\\
\hline

\hspace{-1.8em}\rotatebox{90}{\hspace{-3.7em} \multirow{5}{4.5em}{\bf SIM-4}} 
&\multirow{5}{6.2em}{\includegraphics[width=0.95\linewidth]{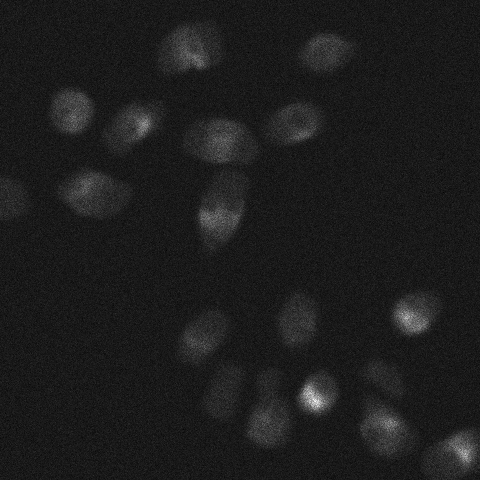}}
& \hspace{-1em} \gmm             & 0.25 & 0.33 & 0.29 & N/A	& N/A  & N/A  & 0.91 & 0.94 & 0.92 & 0.81 & {N/A} & 23\\
& & \hspace{-1em} \kth 			& 0.75 & 0.75 & 0.75 & N/A  & N/A  & N/A  & 0.97 & 0.99 & 0.98 & {\bf0.96} &{{0.98}}  & 5\\
& & \hspace{-1em} \ct 			& 0.75 & 0.60 & 0.67 & 0.75	& 0.68 & 0.72 & 0.86 & 0.97 & 0.92 & N/A & {N/A} & {\bf 1.69}\\
& & \hspace{-1em} \jst 			& 0.75 & 0.38  & 0.50 & N/A	& N/A  & N/A & 0.96 & 0.96 & 0.96 & 0.84 & 0.96 & 4.33 \\
& & \hspace{-1em} \ours 		& {\bf 1.0}  & {\bf 0.80} & {\bf 0.89} & {\bf 1.0}  & {\bf 0.79} & {\bf 0.88} & {\bf 0.98} & {\bf 1.0}  & {\bf 0.99} & {\bf0.96} &\bf{{1.0}}  & 1.85\\
\hline

\end{tabular}
\end{small}
\vspace{-0.2cm}
\caption{Comparison of  our algorithm against state-of-the-art  cell trackers in
  terms  of  tracking  accuracy  and  running time.   It  yields  a  significant
  improvement on  the division and  detection accuracies and performs  either on
  par or slightly better on the migration, MOTA and TRA scores.
  }
\label{tab:results} 
\vspace{-0.5cm}
\end{table*}

  \comment{Note that, we were provided a virtual machine with a precompiled version of the 
  open source \jst{}, which we used to run the experiment and reported the running time.}

%% file: baselines.tex

\subsection{Baselines}

We compared  our algorithm  (\ours) against  the following  four state-of-the-art
methods
\begin{itemize}
    
 \item{\bf  Gaussian Mixture-based  Tracker  (\gmm)~\cite{Amat14}:}  We ran  the
   Gaussian  Mixture Models  approach of~\cite{Amat14},  originally designed  to
   track cell  nuclei, whose code is  publicly available. We manually  tuned its
   parameters to the ones that yield the best results on each sequence.

  \item{\bf KTH  Cell Tracker  (\kth)~\cite{Magnusson12}:} The code  is publicly
    available  and has  been  reported  to perform  best  in  the Cell  Tracking
    Challenge~\cite{Solorzano14,Maska14}.   We   used  the   parameter  settings
    optimized  for  each  dataset  and  provided in  the  software  package.  
   
 \item{\bf  Conservation  Tracking  (\ct  )~\cite{Schiegg13}:}  We  ran  Ilastik
   V1.1.3~\cite{Ilastik11} that implements the method of~\cite{Schiegg13}.  
   We used the default parameters provided with the
   tool to  handle appearance,  disappearance, division and  transition weights.
   The \ct{}  algorithm, like ours,  requires initial segmentations such  as the
   ones shown in the second column of~Fig.\ref{fig:motivation}. We used the same
   segmentations for both  algorithms.  We trained the division  and the segment
   count classifiers separately  for each dataset on  manually labeled
   cells.
   
\item{\bf Joint Segmentation and Tracking  (\jst )~\cite{Schiegg14}:} We ran the
  code  of~\cite{Schiegg14} that  is  publicly available.   The model  comprises
  several  parameters  for oversegmentation  and  tracking,  which we  tuned  to
  achieve the best possible result on each sequence.

\end{itemize}

It is worth noting that, in contrast to all the above state-of-the-art trackers, 
our tracker does not require any user-defined parameters.

To demonstrate the importance of individual  components of our approach, we also
ran simplified versions of \ours{} with various features turned off:
\begin{itemize}

  \item {\bf  Classifier Only (\cl):} We  threshold the output of  our migration
    and division  classifiers at a probability  of 0.5 and return  the resulting
    ellipse detections.

  \item{\bf Best Hierarchy Only (\bh):} For each ellipse hierarchy tree, we only
    keep the level that yields the minimum fitting error, which we define in the
    Supplementary Material.  We then  run our IP  optimization on  the resulting
    graphs, which are smaller than the ones we normally use.

  \item{\bf  Linear  Programming Relaxation  (\lp):}  We  relax the  integrality
    constraint  on the variables  and  solve the  optimization problem  of
    Eq.~\ref{eq:DiscreteOpt5}  using  linear  programming.  We  then  round  the
    resulting  fractional values  to the  nearest  integer to  obtain the  final
    solution.

  \item{\bf  No Conflict  Set  Constraint  (\nc)}: We  remove  the conflict  set
    constraints of Eq.~\ref{eq:exclusion_constraints} and solve the resulting 
    integer program as before.

  \item{\bf Fixed  Division Cost (\fd)}:  We set  the division probability  to a
    constant $p_d$,  which we compute by  finding the relative frequency  of the
    division event in the training sequences.
  
\end{itemize}

%% file: figs/results_ours.tex

\begin{table*}[t]
\centering 
\begin{footnotesize}
\begin{tabular}{l  l || c  c  c | c  c  c | c  c  c | c } 
& & \multicolumn{3}{c}{\bf Division} \vline
& \multicolumn{3}{c}{\bf Detection} \vline
& \multicolumn{3}{c}{\bf Migration} \vline
& \multirow{2}{3em}{\bf MOTA}
\\
&  & Rec & Pre. & F-M. & Rec. & Pre. & F-M. & Rec. & Pre. & F-M. & \\
\hline
\hspace{-1em}\rotatebox{90}{\hspace{-4.5em} \multirow{4}{4em}{\bf HeLa-1}} 
& \hspace{1em} \cl   		& 0.95 & 0.06 & 0.11 & N/A  & N/A  & N/A  & 0.98 & 0.47 & 0.64 & N/A \\
& \hspace{1em} \nc 			& 0.48 & 0.14 & 0.22 & 0.0  & 0.0  & 0.0  & 0.96 & 0.93 & 0.94 & -0.61 \\
& \hspace{1em} \fd			& 0.78 & 0.81 & 0.80 & 0.96 & 0.85 & 0.90 & 0.97 & 0.99 & 0.98 & 0.93 \\
& \hspace{1em} \bh 			& 0.88 & 0.73 & 0.80 & 0.81 & 0.87 & 0.84 & 0.97 & 0.99 & 0.98 & 0.94 \\
& \hspace{1em} \lp 			& 0.92 & 0.80 & 0.86 & 0.95 & 0.81 & 0.87 & 0.97 & 0.99 & 0.98 & 0.93 \\
& \hspace{1em} \ours 		& 0.92 & 0.79 & 0.85 & 0.96 & 0.83 & 0.89 & 0.97 & 0.99 & 0.98 & 0.94 \\
\hline
\hspace{-1em}\rotatebox{90}{\hspace{-4.5em} \multirow{4}{4em}{\bf HeLa-2}} 
& \hspace{1em} \cl   		& 0.91 & 0.10 & 0.18 & N/A  & N/A  & N/A  & 0.92 & 0.60 & 0.72 & N/A\\
& \hspace{1em} \nc 			& 0.73 & 0.31 & 0.43 & 0.06 & 0.02 & 0.02 & 0.95 & 0.96 & 0.95 & 0.35\\
& \hspace{1em} \fd			& 0.77 & 0.82 & 0.80 & 0.84 & 0.78 & 0.81 & 0.96 & 0.98 & 0.97 & 0.90 \\
& \hspace{1em} \bh 			& 0.84 & 0.77 & 0.81 & 0.78 & 0.77 & 0.77 & 0.95 & 0.99 & 0.97 & 0.90\\
& \hspace{1em} \lp 			& 0.85 & 0.83 & 0.84 & 0.84 & 0.78 & 0.81 & 0.95 & 0.99 & 0.97 & 0.90\\
& \hspace{1em} \ours 		& 0.86 & 0.83 & 0.84 & 0.86 & 0.78 & 0.82 & 0.96 & 0.99 & 0.97 & 0.90\\
\hline
\hspace{-1em}\rotatebox{90}{\hspace{-5em} \multirow{4}{4em}{\bf GOWT-2}} 
& \hspace{1em} \cl   		& 0.0  & 0.0  & 0.0 & N/A  & N/A  & N/A  & 0.94 & 0.94 & 0.94  & N/A\\
& \hspace{1em} \nc 			& 1.0  & 0.20 & 0.33 & 1.0  & 0.02 & 0.03 & 0.96 & 1.0 & 0.98 & 0.93\\
& \hspace{1em} \fd			& 1.0  & 0.25 & 0.40 & 1.0  & 1.0  & 1.0  & 0.96 & 1.0 & 0.98 & 0.96\\
& \hspace{1em} \bh 			& 0.0  & N/A  & 0.0  & 1.0  & 1.0  & 1.0  & 0.91 & 1.0 & 0.95 & 0.91\\
& \hspace{1em} \lp 			& 1.0  & 0.50 & 0.67 & 1.0  & 0.50 & 0.67 & 0.96 & 1.0 & 0.98 & 0.96\\
& \hspace{1em} \ours 		& 1.0  & 1.0  & 1.0	 & 1.0  & 1.0  & 1.0  & 0.96 & 1.0 & 0.98 & 0.96\\
\hline
\hspace{-1em}\rotatebox{90}{\hspace{-4.3em} \multirow{4}{4em}{\bf SIM-4}} 
& \hspace{1em} \cl   		& 0.75 & 0.27 & 0.40 & N/A  & N/A  & N/A  & 0.92 & 0.56 & 0.70 & N/A\\
& \hspace{1em} \nc 			& 0.75 & 0.21 & 0.33 & 0.37 & 0.19 & 0.25 & 0.97 & 0.98 & 0.98 & 0.58\\
& \hspace{1em} \fd			& 0.75 & 0.75 & 0.75 & 0.98 & 0.78 & 0.87 & 0.98 & 1.0  & 0.99 & 0.95 \\
& \hspace{1em} \bh 			& 1.0  & 0.80 & 0.89 & 1.0  & 0.78 & 0.87 & 0.98 & 1.0  & 0.99 & 0.96\\
& \hspace{1em} \lp 			& 1.0  & 0.80 & 0.89 & 1.0  & 0.79 & 0.88 & 0.98 & 1.0  & 0.99 & 0.96\\
& \hspace{1em} \ours 		& 1.0  & 0.80 & 0.89 & 1.0  & 0.79 & 0.88 & 0.98 & 1.0  & 0.99 & 0.96\\
\hline

\end{tabular}
\end{footnotesize}
\vspace{-0.2cm}
\caption{Tracking  results with  various features  turned off.   \cl{} does  not
  impose temporal  consistency and  suffers from  low precision.   \nc{} imposes
  temporal  consistency  in the  optimization  but  allows multiple  conflicting
  hypotheses to  appear in the  solution, which  yields a low  precision.  \fd{}
  eliminates competing hypotheses but uses a  fixed cost for the division event.
  \bh{} performs non-maxima suppression on  the hierarchical dimension and hence
  suffers from mis-detection  errors.  
  \lp{} yields fractional flows and the rounding stage eliminates 
  some cell tracks, which leads to a slight drop in performance.
  With all its features  turned on, \ours{}
  achieves the best overall performance. }
\label{tab:results_ours} 
\vspace{-0.4cm}
\end{table*}

%% file: conclusion.tex

\section{Conclusion}
\label{sec:conclusion}

\comment{
We have introduced a novel approach to automatically detecting and tracking cell
populations  in  time-lapse images.   Unlike  earlier  approaches that  rely  on
heuristics to  handle mis-detections  due to clumped  cells and  occlusions, our
approach  simultaneously  tracks cells from an  over-complete  set of  competing
detection hypotheses by solving  a single integer program.  This results  in more accurate
trajectories and improved detection of mitosis events.
}

We have introduced a novel approach to automatically detecting and tracking cell
populations  in  time-lapse images.   Unlike  earlier  approaches that rely either  on
heuristics to  handle mis-detections due to occlusions, or on 
complex integer programs with large sets of variables and constraints,  
our approach yields a simple integer program for simultaneously detecting and 
tracking cells over time. 
Furthermore, we present a robust algorithm for generating over-complete detection 
hypotheses based on fitting ellipses hierarchically. 
This results  in more accurate trajectories and improved detection of mitosis events.

Furthermore, the formalism is very generic. In  future work, we plan to apply it to
people tracking and, in particular, modeling how groups can form and unform.


\comment{Our cell tracker assumes the cell to be more or less ellipse-like 
shape. However
there are cases that a cell is consistently presenting a shape that looks like 
an ensemble of multiple ellipses, where even we human observers have problem 
distinguishing whether it is one cell or more. Unsurprisingly in such cases our
tracker is prone to errors, even though we introduce a feature that accounts for 
the overlap of the multiple cells in this case, where the readers can find more 
details in our supplementary material. To alleviate this problem we may consider
more ellipse-based features based on prior knowledge. This is left an interesting
direction for future work. }